# A New Lagrangian Problem Crossover: A Systematic Review and Meta-Analysis of Crossover Standards

Aso M. Aladdin [1,2,*], Tarik A. Rashid[3]

[1]Department of Information Systems Engineering, Erbil Technical Engineering Colleg, Erbil Polytechnic University, Erbil, KRI, Iraq. aso.dei20@epu.edu.iq
[2]Department of Applied Computer, College of Medical and Applied Sciences, Charmo University, Sulaymaniyah, KRI, Iraq aso.aladdin@charmouniversity.org
[3]Computer Science and Engineering, University of Kurdistan Hewler, Erbil 44001, KRI, Iraq. tarik.ahmed@ukh.edu.krd

**Abstract**

The performance of most evolutionary metaheuristic algorithms relays on various operatives. One of them is the crossover operator, which is divided into two types: application dependent and application independent crossover operators. These standards always help to choose the best-fitted point in the evolutionary algorithm process. High efficiency of crossover operators allows engineers to minimize errors in engineering application optimization while saving time and avoiding costly. There are two crucial objectives behind this paper; at first, it is an overview of crossover standards classification that has been used by researchers for solving engineering operations and problem representation. The second objective of this paper; The significance of novel standard crossover is proposed depending on Lagrangian Dual Function (LDF) to progress the formulation of the Lagrangian Problem Crossover (LPX) as a new standard of systematic operator. The proposed crossover standards result for 100 generations of parent chromosomes are compared to the BX and SBX standards, which are the communal real-coded crossover standards. The accuracy and performance of the proposed standard have evaluated by three unimodal test functions. Besides, the proposed standard results are statistically demonstrated and proved that it has an excessive ability to generate and enhance the novel optimization algorithm compared to BX and SBX.

## 1 Introduction

Combinatorial optimization is stated one of the most widely investigated areas of artificial intelligence. Every year, several research projects focus on issues that arise in this domain. The solution structure, rather than its coding, is utilized by a knowledge-based crossover mechanism for strategic metaheuristic algorithms [1]. Thus, there are several different approaches that researchers have used in the past to solve single or dual problems. However, in the majority of these kinds of applications, the research is limited to solving linear programs, quadratic programs, or, more broadly, convex programming issues. Because the primal optimal solution is closely related to the optimal single or dual solution for convex problems, such a study has assisted investigators in better understanding the relationship [2]. Therewith; a set of optimization algorithms influenced by natural events and animal intelligence is characterized as evolutionary nature-inspired metaheuristic algorithms. They are, thus, frequently nature-inspired algorithms, and samples of these evolutionary metaheuristic algorithms are Genetic Algorithm (GA) [3], Artificial Bee Colony (ABC) [4], Differential Evolution (DE) [5], and with Learner Performance based Behavior algorithm (LPB) [6]. Consequently, popular nature-inspired computing methodsis a field of computer science that could be used and shard with optimization algorithm, computational intelligence, data mining and machine learning [7].

A problem with a fluctuating objective function is one of the more difficult metaheuristic optimization methods, but it is much more common in the real-world search or the self-adaptive optimization. The search technique used to solve such problems for optimality must be flexible enough to adjust to the present function [8]. The The problem with population-based optimizers is that once the investigation has been limited down to the final optimum solution, the diversification population may not be enough to move the search ahead to the new optimum solution. In many of these circumstances, diversity-preserving strategies; such as a high amount of crossover or the use of a clustering operator are required. Therefore; Most of the algorithms have been improved and enhanced in the field of metaheuristic optimization [9]. In addition, there have been several efficient methods presented in this research, which has developed for improving new optimizers depending on crossing genes between parents. Alternatively, new algorithms are always accepted, as long as the suggested methods offer novel improvements or comparable results.

When the genetic recombination operator is chosen correctly, crossing genes can match with the best known techniques for a wide range of problems involving the restrictions. For this determination; it should be suggested or used the superlative crossover standard accomplish the target and the efficiency of this new method might be proved for the feature selection problems [10]. Thus; the motivation behind this paper is that uses conventional techniques and direct search procedures due to the complexity or a large number of variables in the problematic domain. Focusing on some situations, fixing the problem may involve basic algorithm modifications; to give an systematic and meta-analysis study overview; in the innovative metaheuristic optimization. Other motivation side of this paper to propose the novel evolution from several strategies which connected by mathematical assessments with programming methods.

The fundamental goal of this research is to show the basic effectiveness of experience and understanding methods in genetic algorithms. This paper is focused on the generation and standards of crossover and how to affect metaheuristic algorithms. Evidently; crossover, also known as recombination, is a genetic operator in the special process which is exploited to connect the genetic codes of two parents in order to make new offspring (children). Moreover, the crossover technique can be stated as a crucial tool for creating new solutions from a current population stochastically [11]. Many investigations have indicated the influence of the crossover and mutation operators on GAs success. Some of them are concluded success rests in both, whether the crossover is performed alone or mutation alone, or both together. Crossover operators play a substantial role in balancing exploitation and exploration, allowing for features extraction from both chromosomes (parents), with the intention that the developing offspring have beneficial qualities from both chromosomes [12].

For this purpose, LDF [13] is a mechanism that can be used for exchanging genes between chromosomes by locating the replacement chromosome at the greatest or lowest point to improve the new offspring. For example, depending on the meaning of the worst chromosomes for each problem, this point (its worst chromosome) is picked in both chromosomes (parents), or conversely; it can be proposed to the crossover operation between chromosomes to produce the best novel genes for offspring. In addition, this new operator is identified automatically based on performance ratings, and so the time spent selecting the best operator is considered.

The crossover strategy starts with a low value and adjusts it every generation to avoid premature convergence. Besides that, several crossover functions are utilized alternately as a strategy to avoid convergence rate. Nowadays, Population (swarm) algorithms are one of the most successful metaheuristics for managing with these kinds of numerous case problems. As a result, population-based methods have shown to become one of the most efficient methods for dealing with combinatorial function optimization[14]. These techniques operate with several populations of solutions that evolve in tandem with algorithm operation. According this, one of the contributions of this paper is summarizing the previous standards that have generated affording on binary form, real-coded form, and ordered-coded form method. However; it focused on each of them and how the implementation happened to execute the mathematical crossover form. Moreover, the second contribution is proposed the novel method of crossover based on LDF which provide original metaheuristic optimization to build a new optimum solution. The new anticipated LPX is evaluated by comparing it with other previous tuning methods which is a variation on the traditional GA as discussed in section (5).

The next is how this paper is organized: The second part is dedicated to discussing the related work on the crossover types. The third section is stated to illustrate mathematical crossovers standards as a case study overview and some of them are presented by the thinkable pseudocode. The fourth segment is debated the novel formula which can be developed as a standard crossover in the future metaheuristic algorithm is known as LPX The paper also includes a conclusion and proposes the significant feature works. The fifth section is devoted to proving the heuristic crossover result by comparing LPX results with two real-coded crossover standards. Finally, the paper has mentioned the novel feature works.

## 2  Crossover Standards Overview

Many forms of the crossover have been produced over the years and comparisons between various types have been proposed. These started with the one-point crossover and evolved into a range of ways to cover a variety of conditions, particularly uniform crossover[15]. for generating new self-adaptive combinational optimization, a set of assumptions (rules) that are used to simplify natural/biological events, which include a list of control parameters to define the rules of intensification and diversification [16]. The best solution is identified, and other solutions advance toward the best solution according to the given rules. In stochastic mode, the location of a few solutions can be altered and controlled, such as crossover or mutation operation in several metaheuristic algorithms, which is illustrated in a simple flowchart in Figure (1).

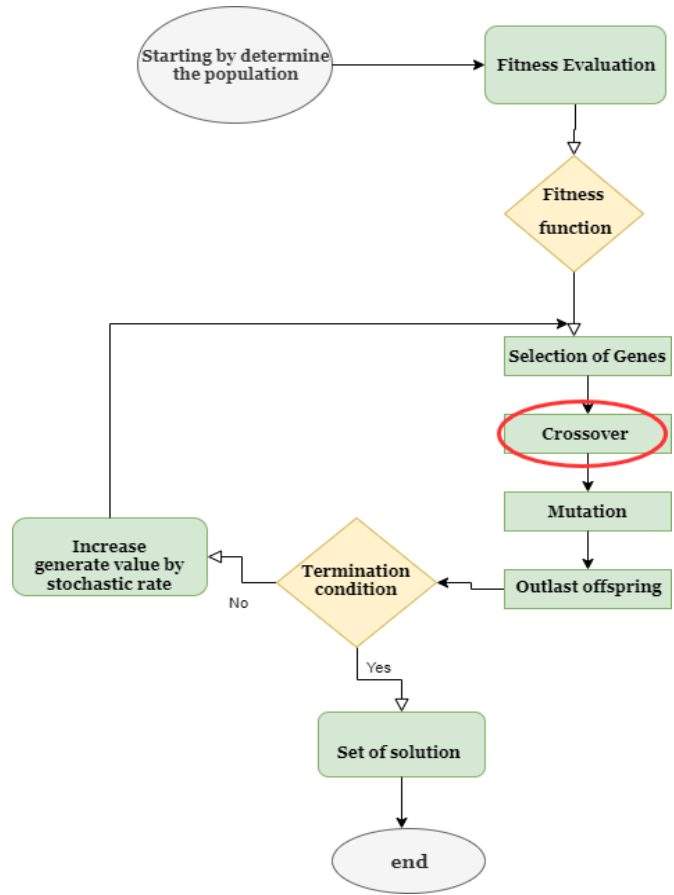

Figure 1: Simple deterministic processes in metaheuristic algorithms

Likewise, several standards for permutation applications, such as the Traveling Salesman Problem (TSP), were defined. For approaching the TSP usage evolutionary algorithms, there exist several representations, such as binary, route, closeness, ordinal, and vector representations. To reduce the overall distance, the researchers presented a novel crossover operator for the TSP [17]. In addition, another new study confirmed sequential constructive crossover (SCX) to fix the TSP in 2010. The primary idea behind this strategy is to choose a random point termed the crossover point, Then, before the crossing point, use a SCX technique with improved edges. After the crossover site, the remaining chromosomes are swapped between parents to generate two children; if a chromosome currently exists, it is replaced with an unoccupied chromosome [18].

Ring Crossover (RC) was offered as a solution to the recombination problem. Parents were grouped in the design of a ring in this type, and then a cut point was picked at random. Parents were grouped in the design of this procedure of circle, and then it is selected a slice point randomly. The other location was the length of the chromosome; the first offspring develops clockwise from the line (original cut), and the second offspring evolves counterclockwiseThey used this type of crossover to aspects mentioned and outperformed the other types of assessed crossover[19].

Despite that, to prevent creating erroneous solutions, evolutionary algorithms that optimize the ordering of a provided huge series require specific crossover operators. It is difficult to point out all of them. Thus, numbers of standard crossovers have already been documented in Table (1) and each of them has been generated for a specific global solution. However, some of them produced the offspring from parents depending on real code; others have relied on the binary coded crossover as stated in the next section.

Table 1: Standard Crossovers Generation

| No. | standard crossover operator name | initial abbreviation | Related work |
|---|---|---|---|
| 1 | Order Crossover Operator | OX1 | [17][24][25] |
| 2 | Order-Based Crossover Operator | OX2 - OBX | [20][24] |
| 3 | Maximal Preservation Crossover | MPX | [20][23] |
| 4 | Alternating Edges Crossover | AEX | [21][23][25] |
| 5 | Edge Recombination Crossover | ERX | [24][25] |
| 6 | Position-Based Crossover Operator | POS | [20][24][26] |
| 7 | Voting Recombination Crossover Operator | VR | [20][24] |
| 8 | Alternating Position Crossover Operator | AP | [24][27] |
| 9 | Automated Operator Selection | AOS | [28] |
| 10 | Complete Sub-tour Exchange Crossover | CSEX | [20][29] |
| 11 | Double Masked Crossover | BMX | [20][30] |
| 12 | Fuzzy Connectives Based Crossover | FCB | [22][31] |
| 13 | Unimodal Normal Distribution Crossover | UNDX | [32] [33] |
| 14 | Discrete Crossover | DC | [34] |
| 15 | Arithmetical Crossover | AC | [19][31][35] |
| 16 | Average Bound Crossover | ABX | [36] |
| 17 | Heuristic Crossover | HC | [17][37] |
| 19 | Parent Centric Crossover | *PCX* | [20][38] |
| 20 | Spin Crossover | SCO | [39] |

In application, crossover standards have typically been classified based on the representation of the gene; genetic sequence has been stored in a chromosome represented by a bit matrix or real code in the different algorithms. Crossover strategies for both techniques are popular, and illustrative instances or classes are genetic recombinations, which are thoroughly explained in the following parts.

## 3  Mathematical Crossover Standards

The exploration for the optimal solution; in metaheuristic algorithms; is mostly based on the generation of new members from current members. The crossover process facilitates the interchange of genetic code between parents, resulting in genes that are more likely to be exceptional to the parents. However, there are so many crossover techniques recorded in the cited study. Researchers should focus on finding and tackling the question of whether the best standard strategy has been improved and adopted.

As mentioned that the crossover operator is comparable to multiplication and biological recombination. In this case,it should be selected more than one genome, and generated one or more offspring for utilizing the genetic codes (Genomes: $G_{x1}$ and $G_{x2}$) as blue balls of the chromosomes (parents) ($X1$ and $X2$) [40] and then produced two new children (Offspring) by two new offspring genes ($O_{x1}$ and $O_{x2}$) as pink balls, this scale operation probability is illustrated in Figure (2). Crossover is typically used in metaheuristic algorithms with a significant probability, particularly in a GA, as challenging in the real-coded [41]. Consequently; the goal of establishing crossover likelihood is to prevent genes loss from the parents, even as offspring could not be better than the parents. According to the distribution of standards of crossover the subsection of 3.1 has been determined the forms of binery crossover. Latterly, the second subsection of 3.2 has classified the catigories of real-coded or floating point crossover. finally; the last subsection 3.3 has devoted for distributing theformof order-coded crossover.

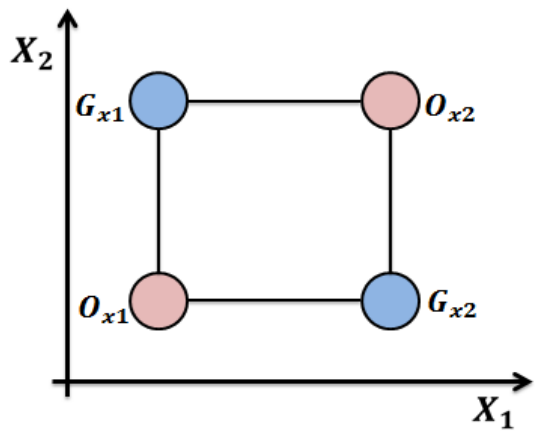

Figure 2: significant probability in the real-coded crossover

### 3.1 Binary Form Crossover

This section would provide a broader collection of crossover operators used in binary-representation for metaheuristic algorithms. The improved previous results presented confirm that most of them are the high-efficiency of crossover according to the implemented problems. As a result, it also could be illustrated how to implement some types of crossover standards and point out some interesting comparisons among others [42]. Traditionally, genetic material has been stored in a gene, which is represented as a bit collection in various techniques. Crossover procedures for bit-order are prominent, and demonstrative instances or categories include genetic manipulation, as explained in the points below.

**Binary Single-point crossover** [43]**:** A crossing point on the parent entity string is picked. Apart from that point in the biological sequence, all data transfer between two units which include biological parents and situational bias is a feature of strings. As indicated in Figure (3), Genes with sub-blocks that including three bits to the right of that point are transferred correspondingly between the two parent. as illustrated by pink and blue color bites and it has generated two new offspring.

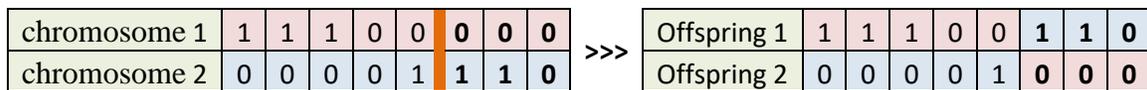

Figure 3: generate new offspring with single-point crossover

**Double-point and n-point crossover** [43]: Randomly generated two locations (strings) or n-point locations on the individual chromosomes are chosen, and gene code is switched at these locations. As seen in Figure (4), two equally spaced points on right and left sides are selected in parent chromosomes then pink color and blue color bites are swapped to perform two single-point crossovers.

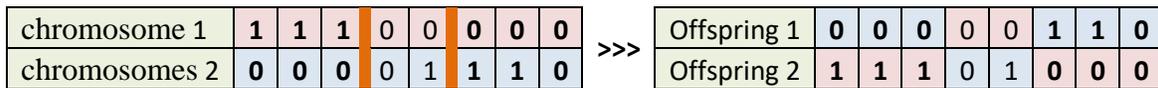

Figure 4: double-points crossover for generating two new children

**Uniform crossover** [44] **and half- Uniform crossover** [45]**:** Like the coin throwing approach, each gene (bit) is chosen at random from one of the comparable genes on the selected parents. Each genome is addressed separately rather than being separated into segments.In this situation, we just flip a coin to see if each genome is present in the child.It may be tossed on supporting of one parent in order to have more genetic information from that parent in the newborn. Figure (5) is illustrated two chromosomes as a two-dimensional array and swapped between them in light blue and pink colors has shown the exchanged bits.

| chromosome 1 | 1 | 1 | 1 | 0 | 0 | 0 | 0 | 0 |
|---|---|---|---|---|---|---|---|---|
| chromosome 2 | 0 | 0 | 0 | 0 | 1 | 1 | 1 | 0 |

>>>

| Offspring 1 | 0 | 1 | 0 | 0 | 1 | 1 | 0 | 0 |
|---|---|---|---|---|---|---|---|---|
| Offspring 2 | 1 | 0 | 1 | 0 | 0 | 0 | 1 | 0 |

Figure 5: producing two fresh offspring by uniform crossover

**Uniform Crossover with Crossover Mask (UCM)** [27]: The matrices are separated into several non-overlapping zones, and the matrix created by the logical operator is known as the crossover mask (CM) generated by the Pseudocode control. Figure (6) presents an example and Pseudocode to show how the new offspring has spawned between chromosome and CM according to the conditions.

```
To generate CM:
if G1= 0 and G2=0
CM =0
if G1!=0 or G2=0
CM = 1
To generate first of gene offspring:
if CM=0  select G1
if CM=1  select G2
To generate gene of the second offspring:
if CM=0   select G2
if CM=1  select G1
```

| chromosomes 1 | **1** | **1** | **1** | 0 | **0** | **0** | 0 | 0 |
|---|---|---|---|---|---|---|---|---|
| chromosomes 2 | **0** | **0** | **0** | 0 | **1** | **1** | 1 | 0 |

| CM | 1 | 1 | 1 | 0 | 1 | 1 | 1 | 0 |
|---|---|---|---|---|---|---|---|---|

| Offspring 1 | 0 | 0 | 0 | 0 | **1** | **1** | 1 | **0** |
|---|---|---|---|---|---|---|---|---|
| Offspring 2 | **1** | **1** | **1** | 0 | **0** | **0** | 0 | 0 |

Figure 6: Pseudocode and example of Uniform Crossover deliveration from Crossover Mask

**Shuffle Crossover (SHX)** [45][46]: initially choose a crossover point at random, such as the highlighted line in Figure (7), then mix the genes code of both parents. It should be emphasized that Shuffle chromosomes for the right and left sites are handled independently. A single point of crossing is chosen and splits the chromosome into two sections known as schema. chromosomes are scrambled in each schema in both parents. To produce offspring, schemas are transferred (as in single crossover)

| chromosome 1 | 1 | 1 | 1 | 0 | 0 | 0 | 0 | 0 |
|---|---|---|---|---|---|---|---|---|
| chromosome 2 | 1 | 0 | 0 | 0 | 1 | 1 | 1 | 0 |

∨∨∨

| Shuffle Occurred Randomly | | | | | | | | |
|---|---|---|---|---|---|---|---|---|
| chromosome 1 | 1 | 1 | 0 | 1 | 0 | 0 | 0 | 0 |
| chromosome 2 | 0 | 1 | 0 | 0 | 0 | 1 | 1 | 1 |

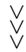

Figure 7: SHX random occurance

**Three-Parent Crossover** (TPX) [47]: According to the prior solution approach, this kind of operator, there are numerous probability rate algorithms to create innovative offspring from three parent genes. Figure (8) highlighted the problems in dealing with new generations with proving by deliberate new offspring according to the general Pseudocode with swapping genes in the elucidation example.

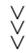

Figure 8: pseudocode and example to explain TPX delebration

### 3.2 Real-coded (Floating Point) Form Crossover

For approaches in which the genes are real-valued and the digits are left alone, speeding up the process without encoding or decoding into binary. Nevertheless, it is less logical than binary representation since crossover has demonstrated that floating-point format can function as well as, if not better than, ordinary binary strings, therefore there is no reason to be concerned about algorithm efficiency if the floating-point encoding is utilized [41]. Several crossover techniques to real-coded were developed. The method is based on effectively adjusted real-coded crossover operations that use the likelihood function to create very distinct sequences that may be candidates for alternative solutions [47]. The crossover techniques have been stated mathematically in the next points.

**Real Single-point Crossover (RSPX)** [48]: It is marginally comparable with a binary single-point crossover; it could be combined with two chromosomes and use a real number for each gene in the crossover point. It could be also generated for two-point, three-point, and n-point crossover. Figure (9) is a specified single-point crossover that includes two genes and swapped the real numbers between them and generates two new offspring.

| chromosome 1 | 0.88 | 0.13 | 0.25 | 0.08 | 0.34 |
|---|---|---|---|---|---|
| chromosome 2 | 0.64 | 0.94 | 0.35 | 0.63 | 0.48 |

>>>

| Offspring 1 | 0.88 | 0.13 | 0.35 | 0.63 | 0.34 |
|---|---|---|---|---|---|
| Offspring 2 | 0.64 | 0.94 | 0.25 | 0.08 | 0.48 |

Figure 9: explain swapping RSPX

**Single arithmetic crossover (SAX)** [47]: Also this standard should be selected a single genome *(k)* randomly in both chromosomes *(n)*. For instance, in Figure (10), *k*= 2 and then define a random parameter (α =0.5). Modify the $k^{th}$ gene of *chromosome1* and *chromosome2* to generate offspring by the selected asthmatic formula that is calculated as equation (1) [49] [50].

$$\text{Gene}_{\text{offspring}(n,k)} = (1 - α) * G_{nk} + α * G_{nk} \qquad (1)$$

$$\text{Gene}_{\text{offspring}(n,2)} = (1 - 0.5) * 0,13 + 0.5 * 0.94 = 0.535$$

| chromosome 1 | 0.88 | 0.13 | 0.25 | 0.08 | 0.34 |
|---|---|---|---|---|---|
| chromosome 2 | 0.64 | 0.94 | 0.35 | 0.63 | 0.48 |

>>>

| Offspring 1 | 0.88 | 0.535 | 0.25 | 0.63 | 0.34 |
|---|---|---|---|---|---|
| Offspring 2 | 0.64 | 0.535 | 0.35 | 0.08 | 0.48 |

Figure 10: target example for SAX

**Whole Arithmetic Crossover (WAX) and Linear Crossover (LX)** [51]: Whole arithmetic (linear) crossover has revelation with a single a similar arithmetic crossover but a calculation handling has happened for all genes from the chromosome $(n)$. The calculation is proved in equation (2) [50] and an example to generate a new offspring has illustrated in Figure (11). Despite that, the probability rates are higher since the number of offspring generated will equal the number of chromosomes (genes = $k$). It should be defined as a random parameter ($α$ and $β$ ) and as ($α_1 = 0.5$ , $α_2 = 1.5, α_3 = -0.5 ... ... α_m$) and ($β_1 = -0.5,$ $β_2 = 0.5, β_3 = 1.5 ... ... β_m$) and for this example when the numbers of genes ($k$) equal three. This example just is calculated and produced three offspring when $k = 1$, in consequence, it can be calculated for *k* numbers however the Figure

(11) just generates three offspring, the generation could be developed other children according to the $k$ numbers.

$$\text{Gene}_{\text{offspring}(m,n,K)} = \alpha_m * G_{nk} + \beta_m * G_{nk} \tag{2}$$

$\text{Gene}_{1,n,1} = \alpha_1 * G_{11} + \beta_1 * G_{21} = \alpha_1 * 0.88 + \beta_1 * 0.64 = \mathbf{0.12}$
$\text{Gene}_{2,n,1} = \alpha_2 * G_{11} + \beta_2 * G_{21} = \alpha_1 * 0.88 + \beta_1 * 0.64 = \mathbf{1.64}$
$\text{Gene}_{3,n,1} = \alpha_3 * G_{11} + \beta_3 * G_{21} = \alpha_1 * 0.88 + \beta_1 * 0.64 = \mathbf{0.52}$

| chromosome 1 | 0.88 | 0.13 | 0.25 |
|---|---|---|---|
| chromosome 2 | 0.64 | 0.94 | 0.35 |

\>\>\>

| Offspring 1 | 0.12 | 0.13 | 0.25 |
|---|---|---|---|
| Offspring 2 | 1.64 | 0.13 | 0.25 |
| Offspring 3 | 0.52 | 0.13 | 0.25 |

Figure 11: Generation of offspring by LX when $k=1$

**Blended Crossover (BX)** [40] [52]: One of the effective crossover that improved several algorithms, If The two-parameter values in pair of chromosomes are G1 and G2, thus G1< G2, then the blend crossover method produces an offspring option in the range *[G1 − α (G2 − G1), G2 − α (G2 − G1)]*. Where $\alpha$ is a constant to be established and so the solutions of the offspring do not exceed the scope of the single variable [40] [53].

The example is stated in Figure (12) that pointed out $k$ number is equal 2, *G1 = 0.13 < G2 = 0.94*, so calculate the range by *[G1 − α (G2 − G1), G2 − α (G2 − G1)]; when α=0.5 the [1.345, 0.535] so,* randomly select *G1 and G2* between the range.

| chromosome 1 | 0.88 | 0.13 | 0.25 |
|---|---|---|---|
| chromosome 2 | 0.64 | 0.94 | 0.35 |

\>\>\>

| Offspring 1 | 0.88 | 1. 2 | 0.25 |
|---|---|---|---|
| Offspring 2 | 0.64 | 0.7 | 0.35 |

Figure 12: BX for second Genes by the range calculation

If the process was evaluated at the previous range, it could not find a global solution as documented in the several improvement algorithms. The condition might be calculated by the new blend formula and found by earlier researchers [40]. Thus, the novel technique has been developed for the BX standard. The parameter γ must be determined with using the $\alpha$ and a random integer $r$ in the range limitation between (0.0, 1.0), both of which are excluded such as the equation (3) [54].

$$\gamma = (1 + 2\alpha) * r - \alpha \tag{3}$$

$Genome_1$ and $Genome_2$ are the offspring solutions which regulate by the parents as equation (4) and (5) [54] consecutively;

$$\text{Gene}_1 = (1 - \gamma) * G_1 + \gamma * G_2 \tag{4}$$

$$\text{Gene}_2 = (1 - \gamma) * G_2 + \gamma * G_1 \tag{5}$$

Figure (13) pointed out the example when $k$ (gene) =2, then identified randomly $\alpha = 0.5$ and $r = 0.5$, so Parameter γ is calculated by equation (3).

$\gamma = (1 + 2\alpha) * r - \alpha = (1 + 2 * 0.5) * 0.5 - 0.5 = \mathbf{0.5}$

$\text{Gene}_1 = (1 - \gamma) * G_1 + \gamma * G_2 = (1 - 0.5) * 0.13 + 0.5 * 0.94 = \mathbf{0.535}$

$\text{Gene}_2 = (1 - \gamma) * G_2 + \gamma * G_1 = (1 - 0.5) * 0.94 + 0.5 * 0.13 = \mathbf{0535}$

| chromosome 1 | 0.88 | 0.13 | 0.25 |
|---|---|---|---|
| chromosome 2 | 0.64 | 0.94 | 0.35 |

\>\>\>

| Offspring 1 | 0.88 | 0.535 | 0.25 |
|---|---|---|---|
| Offspring 2 | 0.64 | 0.535 | 0.35 |

Figure 13: BX for second Genes depending on γ parameter

**Simulated Binary Crossover (SBX)** [55] [56]: it is preferable and common to execute a standard crossover operation among all the standards. As a consequence, SBX functionality has been applied as a standard of the real-coded parameter of GA without any mutation operator and SBX is also created based on the single-point crossover [55]. This approach focuses on the probability distribution of obtainable offspring (Gene) by the specified parents (Genes) as shown in equation (11) [57].

Initially, SBX is calculating the number of children by the formulations (6) and (7) [57] or using formulas (8) and (9) which enhanced the last two formulas by Azevedo [55], which are the most common. The steps to calculate the float number resulting from the crossover are started by fixing a random number μ ~ (0, 1) at first; then calculate the α and generate offspring by using α.

$$\text{Gene}_1 = 0.5[(1 + \alpha_i)G_1 + (1 - \alpha)G_2] \tag{6}$$
$$\text{Gene}_2 = 0.5[(1 + \alpha_i)G_1 + (1 - \alpha)G_2] \tag{7}$$

$$\text{Gene}_1 = 0.5[(G_1 + G_2) - \alpha|G_2 - G_1|] \tag{8}$$
$$\text{Gene}_2 = 0.5[(G_1 + G_2) + \alpha_i|G_2 - G_1|] \tag{9}$$

Subsequently, it must be found the calculation Alpha ($\alpha_i$) [57] is for the two previous offspring as a function of (9). η is the index of a user-defined distribution (not negative) which means η is the number of parameters chosen by the user.

$$\alpha = \begin{cases} (2\mu)^{\frac{1}{\eta+1}}, & if\ \mu < 0.5 \\ (\frac{1}{2(1-\mu)})^{\frac{1}{\eta+1}}, & otherwise \end{cases} \tag{10}$$

Utilize the probability distributions to compute the function of Alpha ($\alpha_i$).

$$f(\alpha) = \begin{cases} 0.5(\eta+1)\alpha^\eta, & \text{if } \alpha \leq 1 \quad \text{(Contracting Crossover)} \\ 0.5(\eta+1)\frac{1}{\alpha^{\eta+2}}, & \text{otherwise} \quad \text{(Expanding Crossover)} \end{cases} \quad (11)$$

When selecting *K2* as a parent 1 and 2 from Figure 14 to get the two new offspring genes, we need to find $(\alpha_i)$ if $\mu = 0.4$ and the user chooses two parameters, the calculation is executed as follows:

$$\alpha = (2*0.4)^{\frac{1}{2+1}} = 0.928$$

$$\text{Gene}_1 = 0.5[(0.13 + 0.94) - 0.928|0.94 - 0.13|] = 0.1592$$

$$\text{Gene}_2 = 0.5[(0.13 + 0.94) + 0.928|0.94 - 0.13|] = 0.9108 \text{ (out of range probability distribution)}$$

| Chromosome 1 | 0.88 | 0.13 | 0.25 |
|---|---|---|---|
| Chromosome 2 | 0.64 | 0.94 | 0.35 |

>>>

| Offspring 1 | 0.88 | 0.1592 | 0.25 |
|---|---|---|---|
| Offspring 2 | 0.64 | 0.9108 | 0.35 |

Figure 14: SBX for the second Genes

However, this technique has more advantages such as; more explorations with a wide range of offspring, and the results are reliable and often end with global optima. But the example exposed that sometimes the result of the offspring gene is out of range according to the probability distributions as shown in Figure (15).

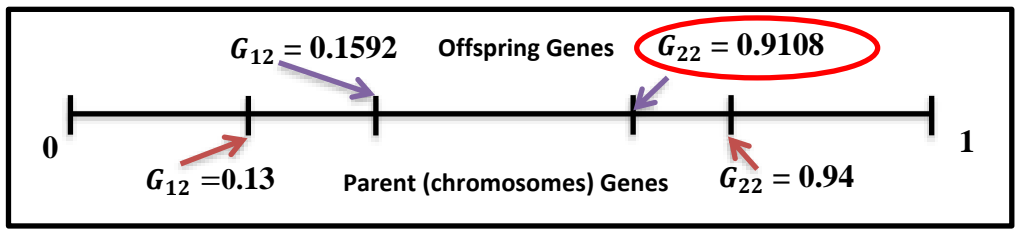

Figure 15: Probability distributions of genes out of range at SBX

### 3.3 Order-coded Problem Methods Crossover

Several various crossover operators have been anticipated by researchers to reproduce either the relative order or the precise organization of chromosomes from the homologous chromosome. The variety of standards-matched crossovers are operators that maintain exact location [58]. As a result, the following sections focus on the most basic forms of order-coded crossovers.

**Partially Mapped Crossover (PMX)** [59]: it produces offspring solutions by transferring a sub-orders from one of the genomes to the other parent while preserving the original sequence (order) with possible several points; determianed the procedure in figure (16). To initiate, choose a crossover range at random and produce children by swapping Genes. Subsequently, in unselected substrings, identify the mapping relationship to permit and legalize the offspring with the mapping connection [60].

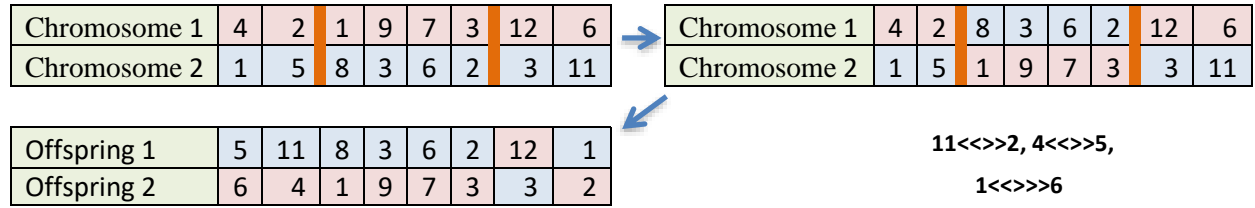

Figure 16: PMX development steps

**The Cycle Crossover Operator (CX)** [17] [61]: determines the number of cycles that exist between two-parent chromosomes. It may be used with numerical strings in which each component appears only once. This assures that each index point in the resultant offspring is filled by a value from one of his parents [62].

According to Figure (17) when the random cycle includes [2, 5, 7, 6, 11], the first offspring world is generated relay on the Pseudocode.

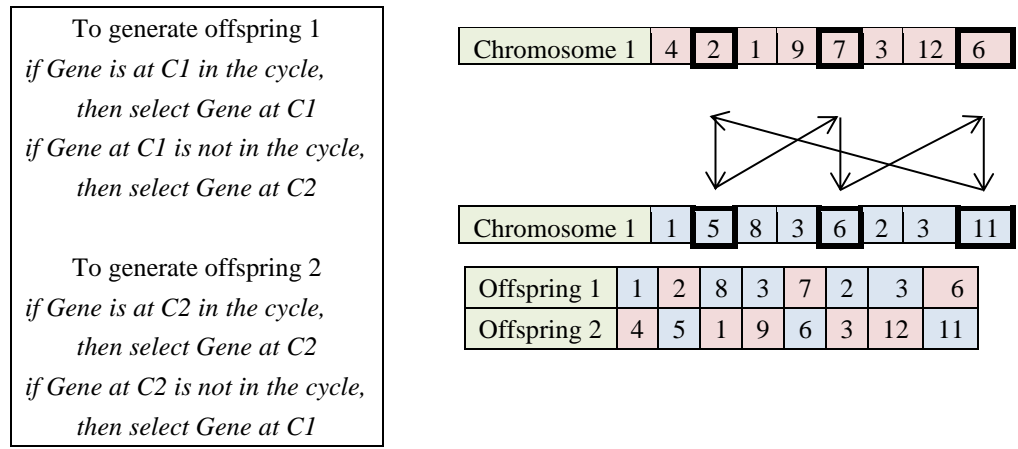

Figure 17: CX Operator Progressive

## 4 Lagrangian Problem Crossover

The major motivation for proving the above crossover standards is to propose and demonstrate a unique substantial crossover technique because Crossover operators will approach those responsibilities with variable degrees of precision. The suggested technique is based on the Lagrange dual function for gene crossings. The Lagrange multiplier approach as a method is used for determining a function of local maxima and local minima with equality constraints or requirements. This issue contains the requirement that one or more equations be exactly solved by the selected variable values [63]. The correlation between the gradient (*slope*) of the function and the gradients (*slopes*) of the constraints leads to a natural new formulation of the global problem, characterized as the Lagrangian function [64] [65]. To calculate the stationary point of $\mathcal{L}$ considered as a function of $\nabla f(x,y) = \lambda \nabla g(x,y)$ and the Lagrange multiplier $\lambda$. Thus, it uses a general method, called the *Lagrange multiplier method* as formulated in equation (12) for solving constrained optimization problems [66] [67].

maximize (or minimize): $(x,y)$ or $f(x,y,z)$

$g(x,y) = c$ If $c$ is a constant
$$\mathcal{L}(x,y,\lambda) = f(x,y) - \lambda\, g(x,y) \tag{12}$$

As illustrated in Figure (18) depicts the restriction $g(x,y) = c$ as a red curve. The blue curves are characteristics $f(x_i, y_i)$. Because of $S1 > S2$, the point where the red constraint tangentially contacts a blue curve is the maximum of $f(x_1, y_2)$ which means tangential with $S1$ in the sideways constraint. Besides, It shows that the assumption that line graphs are tangent has no bearing on the size of any of these gradient vectors, but it is well. When two vectors have the same orientation, we may multiply one by a constant to get the other [68].

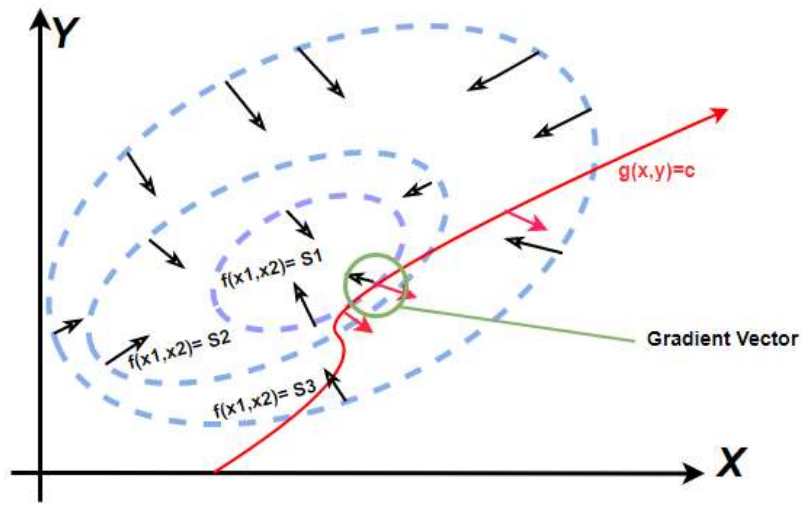

Figure 18: Lagrange multiplier shows the contour lines of two tangent functions when gradient vectors are parallel.

According to the Gradient vectors, this model should be fixed to compute lots of examples to determine the optimum point. It will; for example; assist in physical routing. It will be used to select the smallest point to find the best physical path. Oppositely, it will be used to select the largest point to find the best immune system after the vaccination. However, the Lagrange dual function may be convenient in identifying multiple real global solutions [69].

Consequently, the theorem of LDF depends on real equation examples [70]; it would be proposed for a novel crossover operator and identify several local points. Thus, each station point should be generated a new Gene for offspring from parent (chromosome) chromosomes. Develop an alternative to the Conic Duality theory is called the LDF theory. The Lagrangian Duality Problem theory is more applicable and tangibility for generic nonlinear limitations. [71]. Based on the stationary point in equation (12), the offspring is proposed at equation (13) depending on the LDF theorem problem.

Offspring= $\mathcal{L}(x_1, x_2, \alpha) = f(x_1, x_2) - \alpha\, g(x_1, x_2) = f(x_1, x_2) - \sum_{i=1}^{n=2} \alpha\, g_i(x_1, x_2)$ (13)

For Offspring 1: $\quad f(x_1, x_2) = (x_1 - x_2)^2 + (x_2 - 1)^2$

Subject to $\quad g_1(x_1, x_2) = x_1 + 2x_2 - 1$

$\quad g_2(x_1, x_2) = 2x_1 + x_2 - 1$

For Offspring 2:: $\quad f(x_2, x_1) = (x_2 - x_1)^2 + (x_1 - 1)^2$

Subject to $\quad g_1(x_2, x_1) = x_2 + 2x_1 - 1$

$\quad g_2(x_2, x_1) = 2x_2 + x_1 - 1$

So; we will generate *Offspring1* and *Offspring2* by formulation *offspring1* and *offspring2* equations in the stationary point which includes equation (13) when Lagrangian multiplayer has generated randomly between $-1 \geq \alpha \geq 1$. So, formulate equations (14) and (15);

Offspring1 $= (x_1 - x_2)^2 + (x_2 - 1)^2 - \big(\alpha(x_1 + 2x_2 - 1) + \alpha(2x_1 + x_2 - 1)\big)$ (14)

Offspring2 $= (x_2 - x_1)^2 + (x_1 - 1)^2 - \big(\alpha(x_2 + 2x_1 - 1) + \alpha(2x_2 + x_1 - 1)\big)$ (15)

The form of proposed the standard of crossover; the same as real-coded crossover; creates offspring solutions by inserting a sub-sequence from one of the genomes into the parent while the initial order would be as much point states as reasonable, such as example is calculated by equation (14) and (15) to generate new offspring which has illustrated in figure (19). While $x_1$ is indicated Gene one $G_1$ in the chromosomes one and $x_2$ is indicated Gene two $G_2$, in the chromosomes two and when the stationary multiplier is defined as ($\alpha = 0.2$).

| Chromosome 1 | 0.88 | 0.13 | 0.25 |
|---|---|---|---|
| Chromosome 2 | 0.64 | 0.94 | 0.35 |

>>>

| Offspring 1 | 0.88 | 0.4177 | 0.25 |
|---|---|---|---|
| Offspring 2 | 0.64 | 1.171 | 0.35 |

Figure 19: create two new offspring depending on LPX

## 5   Heuristic Testing Results

Tests achieved a total of 100 stochastic generations (genes) for two parents (chromosomes) with crossover rate values (α) are 0.3, 0.5, and 0.7. However; several standards have been overviewed in the previous sections, but the test has depended on two common real-coded form crossovers which included BX and SBX. These two standard heuristic results are compared by LPX results.

The effective assessment results for both arithmetic crossover and heuristic crossover procedure for addressing metaheuristic algorithm that can demonstrate in this study. Besides; Measures of performance can be based on the sufficient statistical value produced in each iteration. The greater value that calculated according to the average is achieved based on alpha value and it also signifies the superior outcome [72][73][74]. On the standards of crossovers, the outcomes of mathematical calculations from chromosome parent are used to generate a gene on chromosomal offspring [75], where the operation of arithmetic process is found out by depending on the equations (4) and (5) for BX,  the equations (8) and (9)  for SBX, and the equations (14) and (15) for  LPX.

For each gene on chromosome parent approval, the test has depended on three tests of unimodal functions [76] from a group of traditional benchmark functions. The three test equation in the benchmarks for a single variable includes test function one (TF1), test function three (TF3), and test function seven (TF3) as shown in Table 2 [77].  The single result has statistically calculated the summation of generation on chromosome parent and then the average and standard deviation have pointed out to compare all standards relying on alpha value.

Table 2: three unimodal benchmark test functions [77]

| TF | Functions |
|---|---|
| **TF1** | $F(x) = \sum_{i=1}^{n} x_i^2$ |
| **TF3** | $F(x) = \sum_{i=1}^{n} \left( \sum_{j=1}^{i} x_j \right)^2$ |
| **TF7** | $F(X) = \sum_{i=1}^{n} ix_i^4 + roundom\ [0,1]$ |

Convergence and exploitation efficiency have been tested by unimodal benchmark functions test for captivating the algorithm. This number of sample test functions has a single best solution. Nevertheless, these functions are observed as multi-modal test functions, but they have one global optimum. Commonly; In order to reach the global optimum, an algorithm should avoid all local optimal solutions, and these sample test functions might help to conclude the exploration strategy.

Table 3: The performance statistical result test for selected crossover standards with LPX

| | Standards | BX | | | SBX | | | LPX | | |
|---|---|---|---|---|---|---|---|---|---|---|
| α | Test Functions | Sum | Mean | SD | Sum | Mean | SD | Sum | Mean | SD |
| | TF1 | 42.36 | 0.42 | 0.30 | 31.37 | 0.31 | 0.32 | 1737.56 | 17.38 | 17.09 |
| 0.2 | TF3 | 60.00 | 0.60 | 0.66 | 60.00 | 0.60 | 0.66 | 3197.01 | 31.97 | 31.27 |
| | TF7 | 779.24 | 7.79 | 10.78 | 487.58 | 4.88 | 9.52 | **1937510.53** | **19375.11** | **33631.08** |
| | TF1 | 30.00 | 0.30 | 0.30 | 38.58 | 0.39 | 0.30 | 2776.00 | 27.76 | 26.17 |
| 0.5 | TF3 | 60.00 | 0.60 | 0.66 | 60.00 | 0.60 | 0.66 | 5273.89 | 52.74 | 50.06 |
| | TF7 | 461.88 | 4.62 | 9.41 | 661.72 | 6.62 | 10.21 | **4348187.50** | **43481.88** | **64658.48** |
| | TF1 | 35.49 | 0.35 | 0.31 | 46.82 | 0.47 | 0.30 | 3648.64 | 36.49 | 34.78 |
| 0.7 | TF3 | 60.00 | 0.60 | 0.66 | 60.00 | 0.60 | 0.66 | 7019.17 | 70.19 | 67.65 |
| | TF7 | 579.02 | 5.79 | 9.88 | 941.45 | 9.41 | 11.81 | **7300002.73** | **73000.03** | **109185.64** |

Table 3 represents the results of the tests; summation, average, and standard deviation values for the LPX in the three test functions, and the value of alpha is generally greater than BX and SBX. In addition, the higher bold results have shown that the TF7 is high convergence and exploitations for all alpha values. Likewise; LPX has computed as an evolutionary clustering algorithm (ECA) to show the ranking of social class and stochastically analyze meta-heuristic algorithms [78].

BX with SBX in TF1 have determined that the performances are poor nevertheless it has slightly better results for LPX. The gene on chromosome parents for TF1 and three examples alpha value has scattered in figures (20), (21), and (22). Besides, the process of convergence in LPX revealed that the relationships between generations are higher than the other two crossovers standards for all alpha values in the TF3. It proposed that the global optimum solution will be realized in the metaheuristic algorithms and Figures (23), (24), and (25) are illustrated the relationship between generations. Based on TF7 results for all alpha values; LPX has a strangely enormous result as compared to BX and SBX. For more illustration, the gene chromosome comparative results are evaluated and pointed out in Figures (26), (27), and (28).

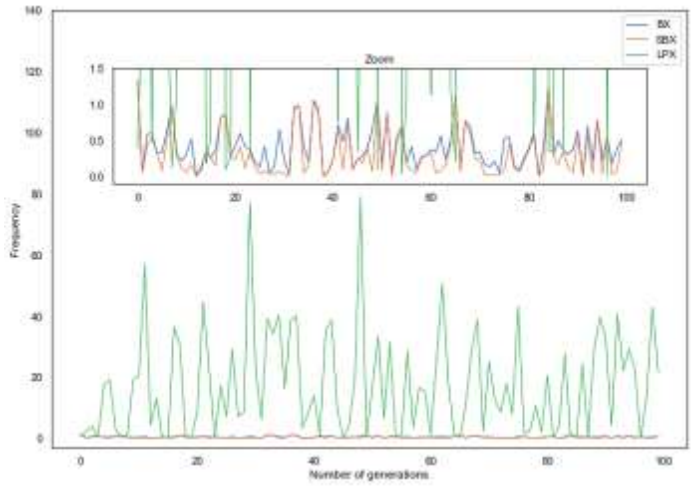
Figure 20: TF1(α=0.2)

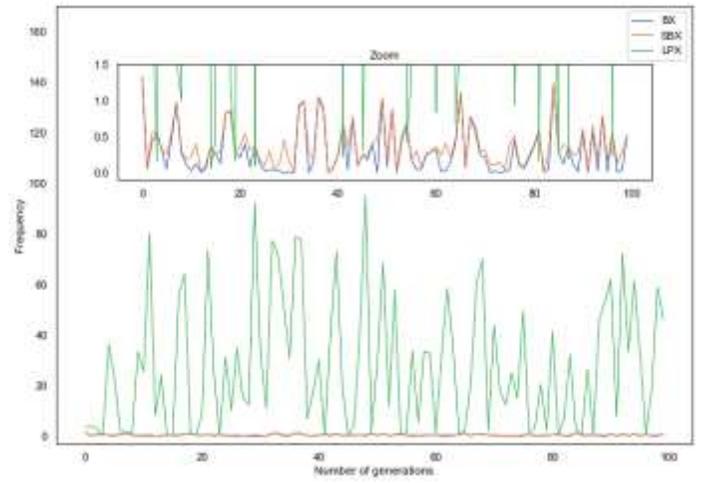
Figure 21: TF1(α=0.5)

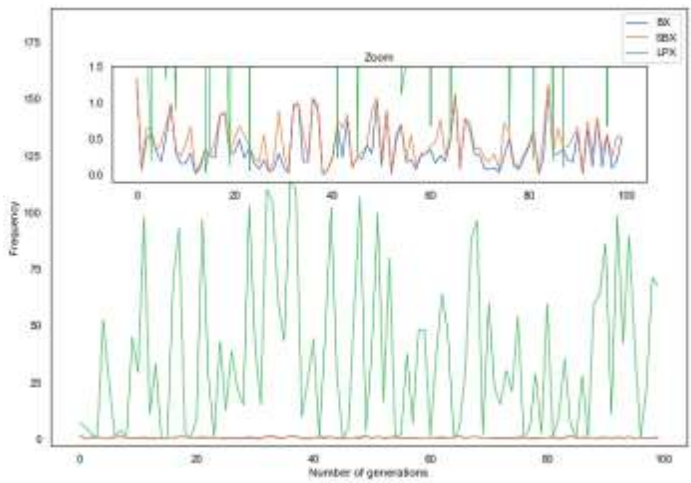
Figure 22: TF1(α=0.7)

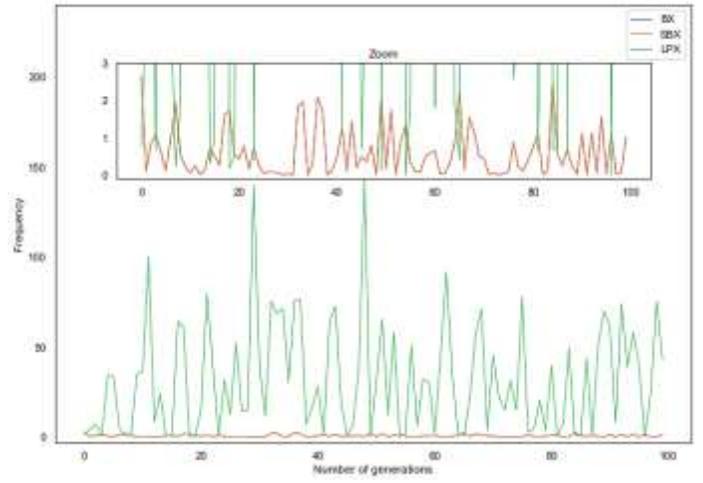
Figure 23: TF3(α=0.2)

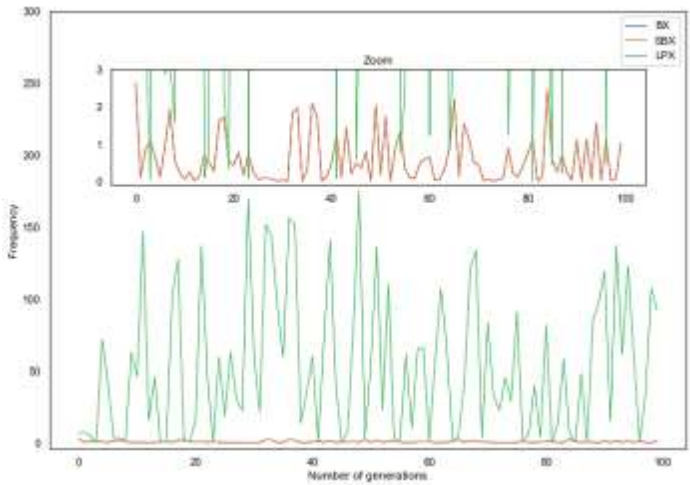
Figure 24: TF3 (α=0.5)

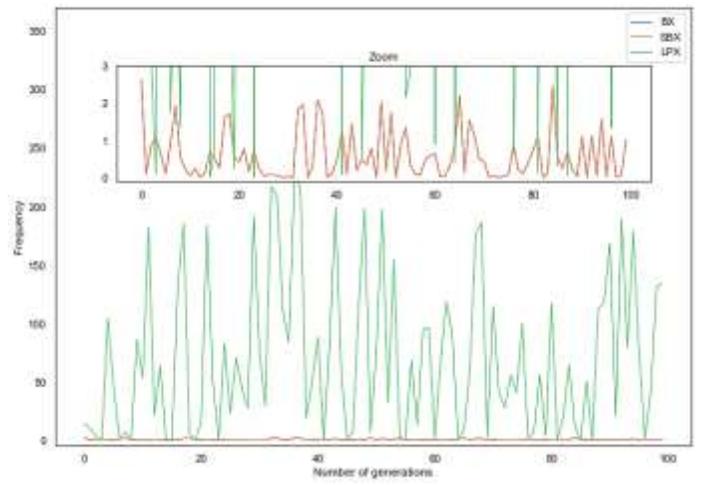
Figure 25: TF3(α=0.7)

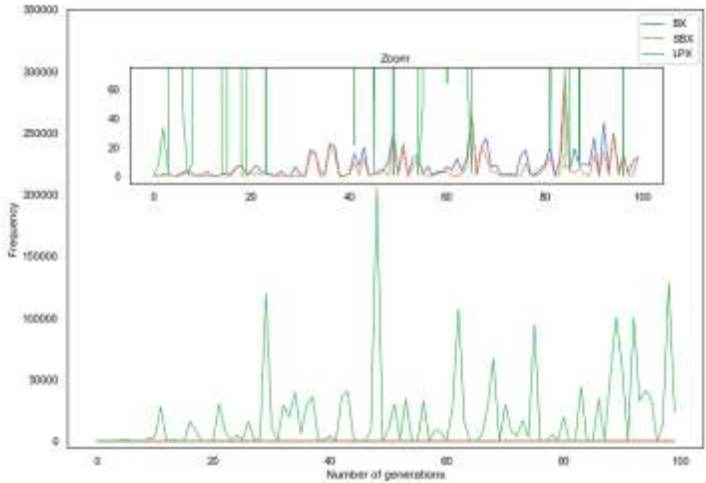
Figure 26: TF7 (α=0.2)

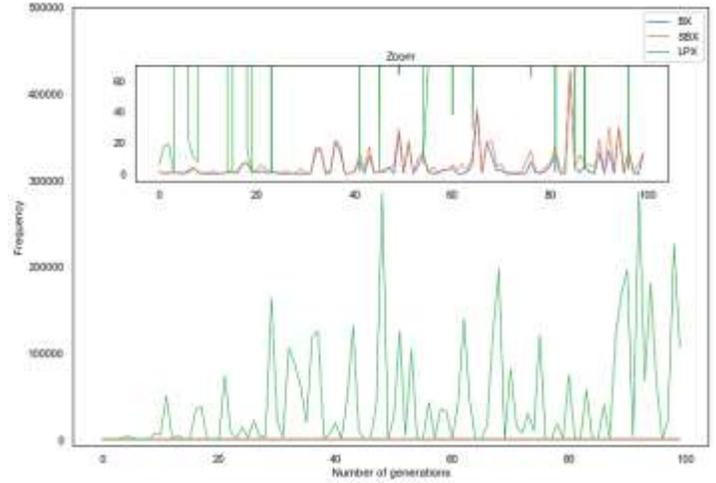
Figure 27: TF7 (α=0.5)

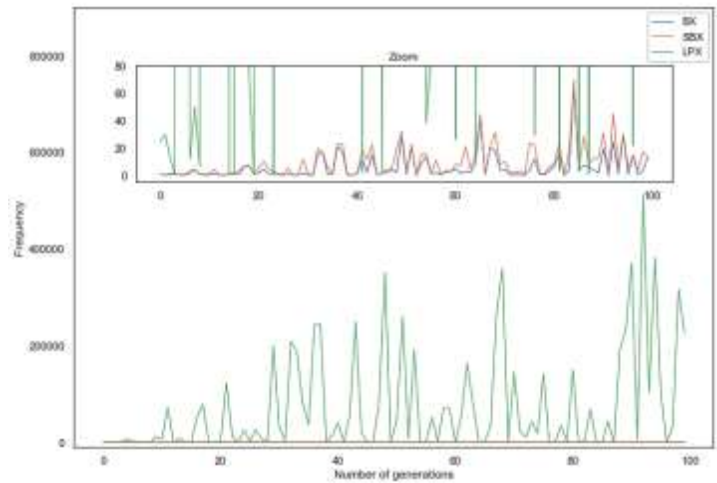
Figure 28: TF7 (α=0.7)

## 6  Conclusion

In conclusion, the most evolutionary metaheuristic algorithm is based on efficient computation techniques that are applied effectively to different problems. Their outcome has been determined by the encoding technique and the selection of standard operators, particularly the crossover operators for enhanced metaheuristic optimization. All of the standards that were described might be adjusted, assisting researchers in obtaining the new crossover operator and selecting a global problem solution. Most of them were easy to compute and hence faster in the calculation, and they may be permitted to produce a wide range of offspring from two parameters as determined by two parent values.

Regardless, dominance may be configured in a variety This work has advocated selecting the optimum crossover standards operator after mathematically reviewing the standard types of crossover operators, such as the binary-coded crossover, real-coded (floating point) crossover, and order-coded problem crossover. In the second part of the research, the novel mathematical approach for crossover standard was spawned that has announced LPX. Additionally, the capability of the technique was tested for generation parent chromosomes compared by BX and SBX. Again, the statistical results of three unimodal test functions proved that the proposed standard had a high convergence and exploration for solving and enhancing new evolutionary algorithms. As a result, it has been suggested that researchers from other disciplines adopt it as a standard technique.

In future works, the researcher can evaluate LPX by compared with other standardized crossover techniques based on binary form, real-code form, or order-coded problem methods crossover. From another perspective, the researcher will improve a novel evolutionary metaheuristic algorithm based on LPX through single-objective optimization or multi-objective optimization. Besides, LPX equations will be improving the frequency-modulated synthesis in the Ant Nesting Algorithm [79] and Child Drawing Development Optimization (CDDO) [80] which they may be gotten the best fitness global solution. Such as; large scale structural optimization modified to evolute several strategies pooled with programming methods as mathematical appraisal [81].